# Using Convolutional Neural Networks to Count Palm Trees in Satellite Images

Eu Koon Cheang(*Author*)
Centre for Computing and Intelligent Systems
Universiti Tunku Abdul Rahman
Kajang, Malaysia
bestceek@1utar.my

Teik Koon Cheang
Centre for Computing and Intelligent Systems
Universiti Tunku Abdul Rahman
Kajang, Malaysia
cheangtk@1utar.my

Yong Haur Tay
Centre for Computing and Intelligent Systems
Universiti Tunku Abdul Rahman
Kajang, Malaysia
tayyh@utar.edu.my

*Abstract*—In this paper we propose a supervised learning system for counting and localizing palm trees in high-resolution, panchromatic satellite imagery (40cm/pixel to 1.5m/pixel). A convolutional neural network classifier trained on a set of palm and no-palm images is applied across a satellite image scene in a sliding window fashion. The resultant confidence map is smoothed with a uniform filter. A non-maximal suppression is applied onto the smoothed confidence map to obtain peaks. Trained with a small dataset of 500 images of size 40x40 cropped from satellite images, the system manages to achieve a tree count accuracy of over 99%.

## I. Introduction

Palm tree localization is mostly done on unmanned aerial vehicle (UAV) images [1][2]. UAVs provide on-demand, high resolution imagery, and are unaffected by cloud coverage. However, UAVs are costly, require operators, and has limited coverage on very large plantations. Satellite imagery on the other hand has much wider coverage than UAV imagery. However the inferior resolution of satellite imagery, the difficulty of obtaining updated satellite imagery and the possible presence of clouds which obscures the palm tree crowns all present a challenge. The task of automated palm tree detection is also further complicated by the fact that palm tree plantations in Malaysia vary wildly in density and spatial arrangement.

Localization is the baseline for further analysis work like yield prediction and estimating fertilizer budget. Palm tree localization using satellite images is not a widely-studied problem. The traditional method is to deploy workers to plantations to manually count trees [8].

## II. Related Work

### A. Palm Tree Mapping

[4] used a novel way to detect palm tree crowns in aerial images with high accuracy. Using a semi-variogram analysis, they were able to detect window size without hand-engineered numbers. However, their approach is only effective on spatially well-arranged palm trees with no overlapping of tree crowns, which is true for the dataset used in their experiment. Many palm plantations in Malaysia have densely planted palm plantations with overlapping canopies. This is especially true for palm plantations with undulating terrain that do not follow a spatial pattern.

## III. Method

We propose a CNN [3], sliding window and image processing approach to the problem.

### A. Classifier

We model the problem as a binary classification task. The CNN is trained with 2 classes of images: a. center of a palm tree crown, and b. does not contain the center of a palm tree crown. We passed these images as input to the CNN and trained it using back propagation to predict the probability that each window image contains the center of a palm tree crown. We compared several CNN models to find the most suitable model for the problem. Lenet [3] was the most suitable, being much less costly to train, need less data, and has a best validation accuracy of 94.5% compared to SqueezeNet [8] at 68.75% and and AlexNet [7] at 67.52%. We suspect that AlexNet and SqueezeNet are overfitting on the training data due to the small dataset.

### B. Sliding Window

The sliding window [5] takes a "window" of fixed size from a larger image, passes it to the classifier, then "slides" to the next step. Given a large image of a plantation, we split it into overlapping segments of 40 x 40 pixels because at the satellite image resolution we are working with, each palm tree crown spans about 40 pixels across. Each 40 x 40 pixel window is then passed to the classifier, which outputs a probability that there is a tree crown centered on that window.

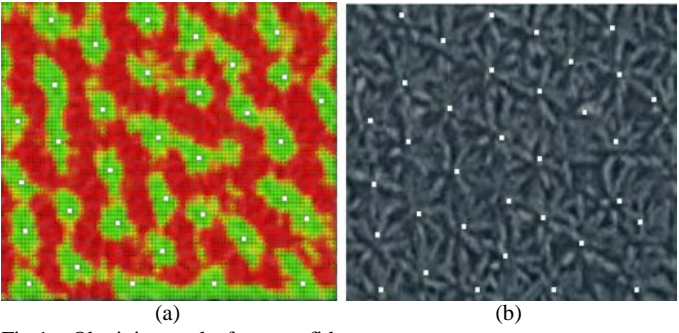

Fig 1. Obtaining peaks from confidence map
(a) confidence map; (b) peaks after filtering

By sliding the window across the entire image, the result is a matrix of classifier confidence (Fig. 1a) given by

$$p(x,y) = c(w(x,y)) \quad (1)$$

where p is the probability that there is a palm tree crown at position (x,y) and c is the classifier's output given input window image w at position (x,y).

*C. Filter*

It is possible for a single palm tree to produce multiple high confidence "peaks" as the sliding window slides over it, especially if the step size for the sliding window is small. There is also noise outputted by the classifier caused by ambiguous window image inputs. These factors present a challenge when performing non-maximal suppression on the output matrix to obtain the local maxima. Therefore we first perform smoothing to remove noise and consolidate multiple peaks before localizing the peaks by applying non-maximal suppression.

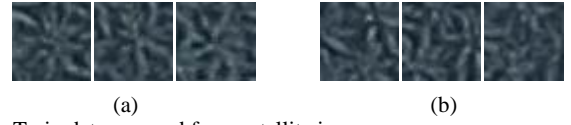

Fig 3. Train data cropped from satellite image
(a) palm trees; (b) not palm trees

## IV. EXPERIMENT

*A. Evaluation Criteria*

The evaluation metric used in this experiment is the margin of error for the number of trees detected after applying non-maximal suppression,

$$margin\ of\ error = \frac{|D-N|}{N} \quad (2)$$

where D is the number of detected trees in the image after applying the filters and N is the actual number of trees in the image.

*B. Dataset*

We made a tool for manually cropping 40x40 window images from Digital Globe satellite imagery and labeled them as one of the two image classes. The palm images are centered on the tree crowns while the non-palm images are randomly chosen windows between tree crowns and from the image background. The dataset consists of 300 images of palm and 500 images of non-palm.

*C. Results*

Our methods were able to obtain results with almost human level accuracy (error margin of < 1%) when applied to images that are similar to the training images.

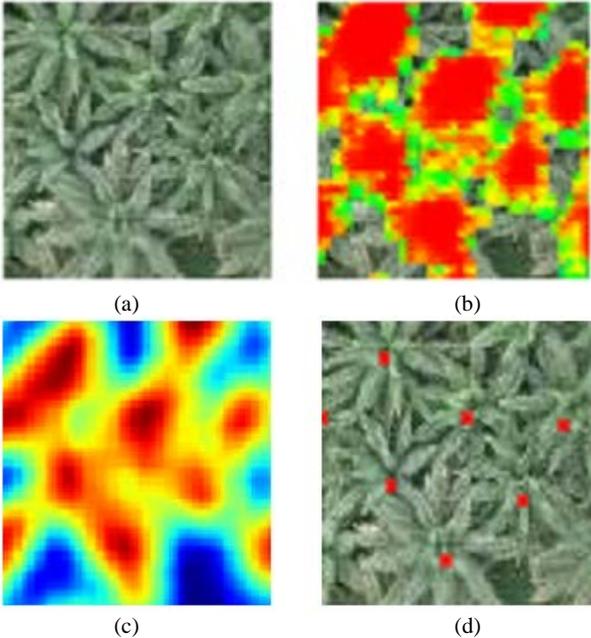

Fig 2. Localization process
(a) original image; (b) confidence matrix overlaid over original image; (c) confidence matrix after smoothing; (d) detected peaks after non-maximal suppression

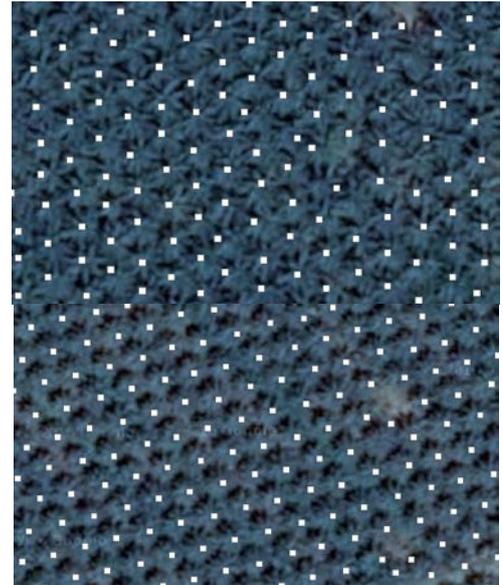

Fig 4. Localized trees in satellite imagery
(a) matured trees with overlapping canopies; (b) adolescent trees

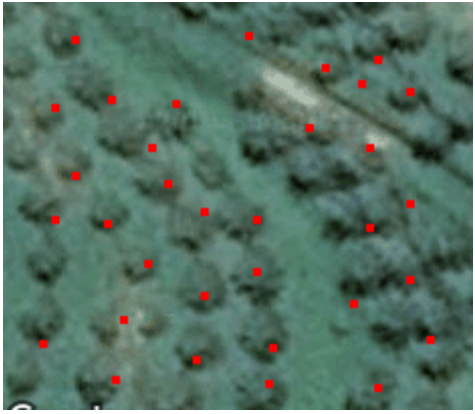

Fig 5. Detected peaks in different satellite imagery

When our approach is applied to drastically different satellite images like the one in Fig. 5, the performance is poor. This is due to the small dataset which only contains images similar to that of Fig 4. Rudimentary experiments show that our approach can adapt to this new type of image after training the CNN on the new image type. This is better than hand-crafted feature engineering because there is no need to reengineer the feature extraction process.

## V. CONCLUSION

Through usage of our proposed method, we were able to significantly reduce the time and effort in counting and localizing palm trees with human-level accuracy. However, the counting of palm trees in satellite images is still a challenge compared to UAV because of poor image resolution, cloud coverage and image availability. Although the trained CNN model is also not robust to different types of images, it responds well to new training data and can quickly adapt.

Future work can include multispectral data as additional dimensions of classifier input instead of only using the red, green and blue spectrums as input. This gives more context to the input images especially as the infrared and near-infrared channels has shown strong correlation to the presence of vegetation.

REFERENCES

[1] Malek, S., Bazi, Y., Alajlan, N., AlHichri, H. and Melgani, F., 2014. Efficient framework for palm tree detection in UAV images. IEEE Journal of Selected Topics in Applied Earth Observations and Remote Sensing, 7(12), pp.4692-4703.

[2] Aliero, M.M., The Usefulness Of Unmanned Airborne Vehicle (UAV) Imagery For Automated Palm Oil Tree Counting. [Online]. Available: https://www.researchgate.net/publication/281939505_The_Usefulness_Of_Unmanned_Airborne_Vehicle_UAV_Imagery_For_Automated_Palm_Oil_Tree_Counting

[3] LeCun, Y., Bottou, L., Bengio, Y. and Haffner, P., 1998. Gradient-based learning applied to document recognition. Proceedings of the IEEE, 86(11), pp.2278-2324.

[4] P. Srestasathiern and P. Rakwatin, 2014. Oil palm tree detection with high resolution multi-spectral satellite imagery. *Remote Sensing.* [pdf] Available: http://www.mdpi.com/2072-4292/6/10/9749/pdf

[5] Wang, T., J.Wu, D., Coates, A. and Y. Ng, A., 2012. End-to-end text recognition with convolutional neural networks. In: *Pattern Recognition (ICPR).* [online] IEEE, pp.3304-3308. Available: http://ai.stanford.edu/~ang/papers/ICPR12-TextRecognitionConvNeuralNets.pdf

[6] Forrest N. Iandola , Matthew W. Moskewicz , Khalid Ashraf , Song Han , William J. Dally and Kurt Keutzer, 2016. *SqueezeNet: AlexNet-level accuracy with 50x fewer parameters and <1 MB model* size. [pdf] Available: http://arxiv.org/pdf/1602.07360v2.pdf

[7] Krizhevsky, A., Sutskever, I. and Hinton, G.E., 2012. Imagenet classification with deep convolutional neural networks. In *Advances in neural information processing systems* (pp. 1097-1105).

[8] Chen, Z.Y., 2015. Capturs: An Automatic Palm Tree Counting System. [pdf] Available:http://www.aarsb.com.my/wp-content/Publication/Newsletter/PDF/2015-Oct.pdf